# A hierarchical residual network with compact triplet-center loss for sketch recognition

Lei wang[1,2] . Shihui Zhang[1,2] . Huan He[3] . Xiaoxiao Zhang[1,2] . Yu Sang[1,2]

**Abstract**
With the widespread use of touch-screen devices, it is more and more convenient for people to draw sketches on screen. This results in the demand for automatically understanding the sketches. Thus, the sketch recognition task becomes more significant than before. To accomplish this task, it is necessary to solve the critical issue of improving the distinction of the sketch features. To this end, we have made efforts in three aspects. First, a novel multi-scale residual block is designed. Compared with the conventional basic residual block, it can better perceive multi-scale information and reduce the number of parameters during training. Second, a hierarchical residual structure is built by stacking multi-scale residual blocks in a specific way. In contrast with the single-level residual structure, the learned features from this structure are more sufficient. Last but not least, the compact triplet-center loss is proposed specifically for the sketch recognition task. It can solve the problem that the triplet-center loss does not fully consider too large intra-class space and too small inter-class space in sketch field. By studying the above modules, a hierarchical residual network as a whole is proposed for sketch recognition and evaluated on Tu-Berlin benchmark thoroughly. The experimental results show that the proposed network outperforms most of baseline methods and it is excellent among non-sequential models at present.

**Keywords** sketch recognition . residual learning . multi-scale fusion . compact triplet-center loss

## 1 Introduction

Sketch is a simple yet efficient communication tool which can describe objects and scenes that are difficult to be expressed in words. And the widespread use of touch-screen devices makes this tool more common. Therefore, more and more researchers begin to focus on the related research about sketch, including sketch recognition [1, 2], forensic sketch analysis [3, 4], sketch-based image retrieval [5-7] and sketch-based 3D image retrieval [8, 9]. Among them, sketch recognition is a very challenging task. The reasons are that: (i) sketches are highly abstract and iconic. They do not contain rich colors and textures but only a few simple strokes, which leads to the wonderful similarity between some different sketches. (ii) There are many styles to draw the same kind of sketches, and different people

___________________________________________________

✉ Shihui Zhang

    sshhzz@ysu.edu.cn

1. School of Information Science and Engineering, Yanshan University, Qinhuangdao 066004 Hebei, China
2. Key Laboratory for Computer Virtual Technology and System Integration of Hebei Province, Qinhuangdao 066004 Hebei, China
3. School of Mathematics and Information Science and Technology, Hebei Normal University of Science and Technology, Qinhuangdao, 066004, China draw the same kind of sketches at different drawing levels, which makes sketches from the same class look different. Tu-Berlin [10], a large-scale sketch dataset shows that sketches from the same class present different appearances while sketches from different classes have similar appearances. Fig. 1 shows some examples. Accordingly, it is critical to improve the distinction of the sketch features from different classes and minimize intra-class difference.

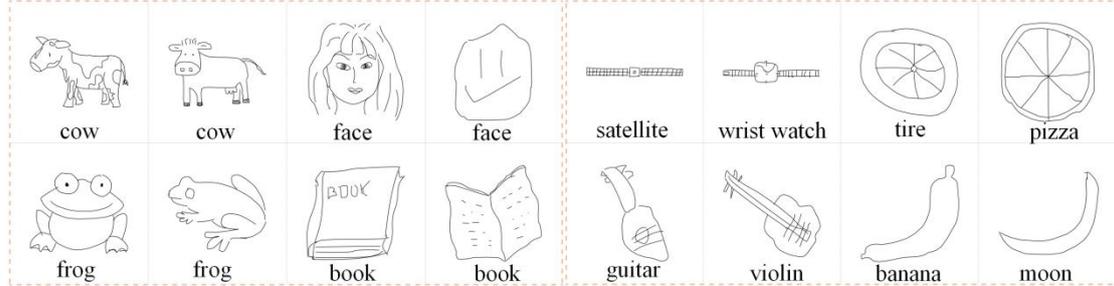

**Fig. 1** Sketches of same class differ in the left figure and sketches of different classes look similar in the right figure

Prior works include the hand-crafted methods [10-12] and the deep learning methods [1, 2, 13, 14, 15]. The former is to extract the hand-crafted features followed by applying an aggregated method (such as bags-of-words) to generate the final representation. Then the representation is fed to a classifier for recognition. The latter designs neural networks using sketch characteristics such as stroke sequences, sketch sparsity or textures of related images so as to realize the purpose of sketch recognition. To improve the distinction of the sketch features, most deep learning methods exploit additional information or operations such as sequential strokes [2, 15, 16, 17], specific data augmentation operation [1, 18], multiple training process[1, 2, 13] and so on. These methods either have the complicated training processes or place higher requirements on the training samples. And our method improves the distinction of the sketch features from three aspects. Specifically, we make contributions as follows: (i) A specific residual block is designed to learn the complementary information from multiple scales, which enhances the distinction of sketch features from different classes. (ii) From the perspective of network structure, the network is a recursive residual block with three levels. In contrast with ResNet that has only several single-level residual blocks, the network can better handle with the degradation problem [19], and exploit the structure with three levels to enhance its own discriminative power efficiently. (iii) We propose a special loss, the compact triplet-center loss, which can make the sketch features from different classes separate as much as possible in sketch feature space. The above three contributions can make the sketch features highly distinguishable, which results in the high recognition accuracy.

The proposed network can avoid the dependence on sequential information because it is based on CNN instead of a sequential model. Meanwhile, due to the simple structure of our network, it only takes one training phrase with a simple training method. In a word, we propose a hierarchical residual network with compact triplet-center loss specifically for sketch recognition. And the network has a simple training method and low requirements for the training samples, which makes it present more practical significance. We conduct numerous comparative experiments on Tu-Berlin benchmark, and the results show that the performance of our network is better than most of the baseline methods.

The remainder of paper is organized as follows. Section 2 briefly introduces related work. Section

3 describes our main methods for sketch recognition. Comparative experiments and ablation analysis are presented in Section 4. Finally, we summarize the article.

Important abbreviations used in this paper are summarized in Table 1.

**Table 1** The abbreviated names and their corresponding full names.

| Abbreviated name | Full name |
|:---:|:---:|
| chi2 | chi square |
| hi | histogram intersection |
| CNN | convolutional neural network |
| DVSF | deep visual-sequential fusion |
| HOG | histogram of oriented gradient |
| LSTM | long short-term memory |
| RBF | radial basis function |
| RNN | recurrent neural network |
| SSIM | self-similarity |
| SVM | support vector machine |

**2 Related work**

**2.1 Sketch recognition**

Inspired by image recognition methods, the early sketch recognition methods are mainly realized by hand-crafted features. Eitz et al. [10] combine HOG with bags-of-words model to classify sketches with 56% accuracy using multi-class support vector machine. Similarly, Schneider et al. [11] extract dense sift descriptor, then combine the descriptor with Fisher Vector to obtain 68.9% accuracy. Li et al. [12] use RBF kernel, linear kernel, chi2 kernel and hi kernel for multi-kernal learning on HOG, SSIM, Daisy and Start graph respectively. Although the accuracy of this method is only 65.81%, it has lower memory consumption than Fisher Vector [11]. The hand-crafted feature methods have achieved some improvements in accuracy. However, due to the limited representation of the hand-crafted features and the lack consideration for variation in appearance, there is still a gap between the recognition ability of these methods and that of humans. In recent years, with the rise of researches related to neural network [20-22] and the release of a large-scale sketch dataset [10], it is possible to accomplish the sketch recognition task by deep learning. The studies can be roughly divided into two categories. One is based on sequential models such as RNN, LSTM, and a sketch is regarded as sequential strokes. Accordingly, the sketch features usually rely on the temporal sequences in datasets, but many datasets do not meet this condition. Related papers include Sketch-BERT [2], $S^3$Net [15], SketchMate [16] and DVSF [17]. The other is based on CNN. This kind of method treats a sketch as a static image, following the basic method of image classification, to design the network architecture according to characteristics of the sketch. The related researches include Sketch-a-Net [1], SketchNet [13] and transfer learning [14]. Generally, the recognition accuracy of sequential models is higher than that of CNN. However, to avoid the dependence on sequential strokes, this paper makes a profound study on the sketch recognition network based on CNN. It is exciting that our network based on CNN can achieve comparable performance with the best sequential model.

**2.2 Multi-scale fusion**

Multi-scale fusion is widely used in the field of computer vision such as image classification [23, 24], salient object detection [25, 26], scene parsing [27]. Zhang et al. [23] extract multi-scale features based on a stacked sparse autoencoder, and a new pooling strategy is proposed to reduce dimensionality for these features. Their experiments verify that the method can remain the details in images. Pang et al. [24] apply the transformation-interaction-fusion strategy to the self-interaction module. By embedding the module into each decoder unit, the network can adaptively extract multi-scale information, so as to cope with the multi-scale variation of significant objects. PspNet [27] uses a pyramid pooling module with four different scales so that the network has the ability to obtain global context. From the above researches, it is concluded that multi-scale fusion can effectively improve the ability to distinguish the sketch features by increasing the perception for multiple scale information. Yu et al. [1] apply multi-scale fusion to the sketch recognition task. Unfortunately, their model is not an end-to-end network and the parameters are increased obviously to fuse four scale features. GoogLeNet [22] is the first convolutional neural network to realize multi-scale feature fusion in building blocks where multi-scale features can be fused by using shortcut connections. Motivated by the above works and facts, we put multi-scale fusion into building blocks instead of a network structure. And we propose a multi-scale residual block which can easily realize muti-scale fusion with proper increase in the number of parameters.

**2.3 Residual learning**

The idea of residual learning firstly appeared in ResNet [19], which was used to solve the vanishing gradient caused by the increased depth in traditional convolutional neural networks. Residual learning is widely used in computer vision tasks such as image classification [19, 28, 29], haze removal [30], video person re-identification [31], etc. Generally, residual learning can improve the performance of the network for computer vision task. In ResNet [19], the identity mapping is used to build a basic residual block. The network depth is increased to 152 layers by stacking multiple basic residual blocks. Compared with the plain network, the residual learning method has lower training error and higher recognition accuracy. Wang et al. [32] stack multiple bottom-up and top-down forward attention modules, and use residual learning to solve the gradient vanishment caused by stacking multiple attention modules, which enables different attention modules to be fully learned. Yeh et al. [30] use a simplified U-Net network to remove haze in fogged images, and exploit residual learning to realize multi-scale fusion so as to solve the color distortion problem. In the above studies, the idea of residual learning is basically reflected in the residual block. However, we believe that it is more beneficial to the network performance by integrating the residual learning idea into both the local residual block and the overall network structure. Therefore, this paper proposes a hierarchical residual network to achieve all-round fusion from local to global.

**2.4 Some important losses**

Loss functions are of vital importance in guiding the training of neural networks. Among them, softmax loss is the most common and plays an important role in sketch classification [1, 13], face recognition [33, 34], object detection [35] and so on. Nevertheless, softmax loss is only used for basic classification which does not consider the sparsity of intra-class space and the compactness of inter-class space. Consequently, He et al. [36] develop triplet-center loss and combine it with softmax loss to train their network in 3D shape field. The triplet-center loss's definition is as follows.

$$L_{tc} = \sum_{i=1}^{M} \max(D(f_i, c_{y^i}) + m - \min_{j \neq y^i} D(f_i, c_j), 0) \qquad (1)$$

where $f_i$ is the feature of a input sample, $c_{y^i}$ indicates the corresponding center of $f_i$, $c_j$ is a negative center (the corresponding center of a negative input sample), and *m* is a margin which is enforced between two different classes. The triplet-center loss uses the nearest negative center to keep the input sample's center away from the other classes in each iteration. Since the joint loss can better alleviate large variances from identical classes and few variances from different classes, we hope to introduce this loss into the sketch recognition task. However, the triplet-center loss doesn't work well because it is specifically designed for 3D shape retrieval task and doesn't take specific properties of sketch into account. Therefore, a compact triplet-center loss is proposed to solve the aforementioned problem.

## 3 Methodology

### 3.1 Overview of the proposed method

To accomplish the sketch recognition task, we construct a hierarchical residual network with compact triplet-center loss. The overall architecture of our proposed network is shown in Fig. 2. It is mainly composed of three parts including a front-end network, a hierarchical residual structure and a classification module. The front-end network takes a sketch as input for extracting shallow features, then the hierarchical residual structure is used to sufficiently learn the final sketch features. Finally, the classification module exploits these features to classify for the recognition result.

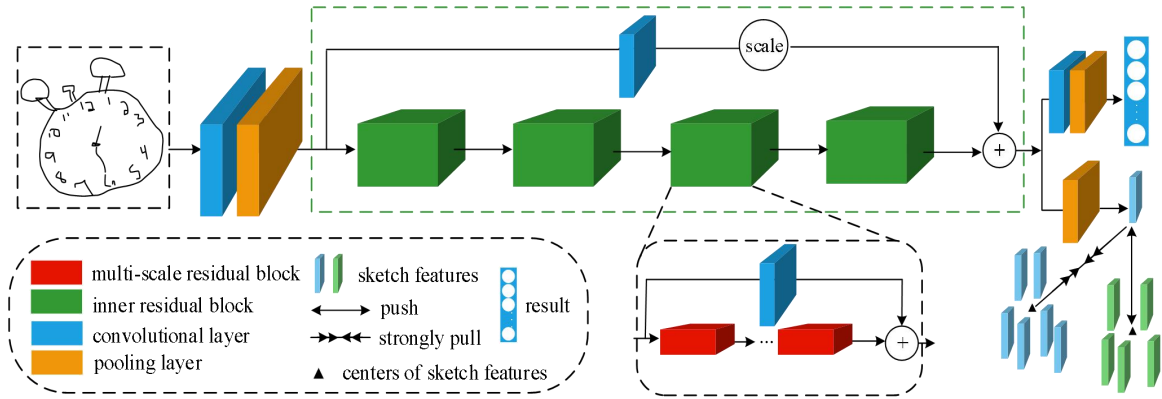

**Fig. 2** The overall architecture of hierarchical residual network with compact triplet-center loss

Specifically, we assume that $S$ is a sketch as the input for our network. $S$ is fed into a front-end network consisted of a convolutional layer and a max-pooling layer to extract shallow features. Given a sketch $S$, the process can be expressed as follows.

$$f_s = F_p(F_c(S)) \quad (2)$$

where $f_s$ represents the shallow features, $F_p$ and $F_c$ denote the max-pooling layer and the convolutional layer respectively.

To enhance the discriminative power of the sketch features, we construct a hierarchical residual structure by stacking multi-scale residual blocks and using shortcut connections. The network can employ the structure to better handle with the vanishing gradient and to fully learn the sketch features. For simplicity, we assume that every level only contains one residual block. Given a shallow feature $f_s$, the output of the current structure can be represented as follows.

$$f_h = F_o(F_i(F_m(f_s) + f_s) + f_s) + \beta f_s \quad (3)$$

where $F_m$ denotes the nonlinear mapping function of the multi-scale residual block, $F_i$ denotes the inner nonlinear mapping function of the inner residual block, $F_o$ denotes the outer nonlinear mapping function of the outer residual block and $\beta$ denotes a scaling factor. The hierarchical residual structure is shown in the green dotted box of Fig. 2.

**Table 2** The architecture of the proposed network. Convolutional or pooling layer can be written as "kernal size×kernal size, channel numbers, stride, operation".

| Module name | | Layer | Output size | Parameters and operations |
|---|---|---|---|---|
| Front-end network | | 1 | 112×112×64 | 7×7, 64, 2, conv |
| | | 2 | 56×56×64 | 3×3, 2, max-pooling |
| Hierarchical residual structure | Outer residual block | Inner residual block 1 | 1 | 56×56×64 | $\left\langle \begin{matrix} 3\times3,64 \\ 3\times3,64 \end{matrix} \middle| 3\times3,64 \right\rangle \times 3$ |
| | | Inner residual block 2 | 2 | 28×28×128 | $\left\langle \begin{matrix} 3\times3,128 \\ 3\times3,128 \end{matrix} \middle| 3\times3,128 \right\rangle \times 4$ |
| | | Inner residual block 3 | 3 | 14×14×256 | $\left\langle \begin{matrix} 3\times3,256 \\ 3\times3,256 \end{matrix} \middle| 3\times3,256 \right\rangle \times 6$ |
| | | Inner residual block 4 | 4 | 7×7×512 | $\left\langle \begin{matrix} 3\times3,512 \\ 3\times3,512 \end{matrix} \middle| 3\times3,512 \right\rangle \times 3$ |
| Classification module | | Branch 1 | 1 | 7×7×250 | 1×1, 250, 1, conv |
| | | Branch 1 | 2 | 250 | 7×7, 7, average pooling |
| | | Branch 2 | 1 | 512 | 7×7, 7, average pooling |

The classification module consists of two branches. The first branch includes a convolutional layer and an average pooling layer for dimensionality reduction and a softmax function for main classification. The second branch includes an average pooling layer for dimensionality reduction and a compact triplet-center loss for strengthening distinguishing ability. Given a hierarchical feature $f_h$, the recognition result can be written as follows.

$$c = \psi_s(F_p F_c(f_h)) \tag{4}$$

where $F_p$, $F_c$ and $\psi_s$ stand for an average pooling layer, a convolutional layer and a softmax classifier respectively. In addition, the compact triplet-center loss is an auxiliary function during training. The softmax loss combines with it to achieve more discriminative sketch features. The joint loss can be formulated as follows.

$$L_j = \psi_s(F_p F_c(f_h)) + \lambda \psi_c(F_p(f_h)) \tag{5}$$

where $L_j$ represents the joint loss, $\psi_c$ represents the compact triplet-center loss. The combination of the softmax loss and the compact triplet-center loss is exploited to supervise the training process of the overall network.

Details of the three parts are shown in Table 2. The front-end network represents a combination of a convolutional layer and a pooling layer. It is the first two layers in Fig. 2. In Table 2, the hierarchical residual structure only shows the parameters of the multi-scale residual blocks. Details of the inner and outer residual blocks can refer to Subsection 3.3. The classification module corresponds to the end of the network in Fig. 2.

### 3.2 Multi-scale residual block

Multi-scale fusion is a common method in the image classification field because it is consistent with the human visual characteristics. For a natural image, an object usually presents a profile or shape when human eyes are far away from the object, while the object usually shows texture, structure and other details when human eyes are near it. Multi-scale fusion integrates global information and details to deepen the understanding about image features so as to achieve the discriminative sketch features. Similarly, sketches also need multi-scale fusion. Although some details are removed from sketches, many sketches still contain regions with different scales as follows.

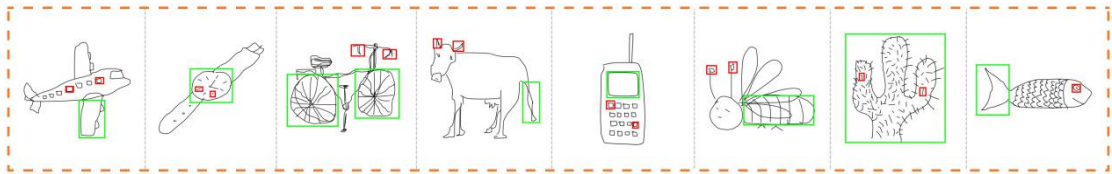

**Fig. 3** Some sketches with multiple scale regions. Different color boxes indicate different scales. Thus, increasing the types of scales is beneficial to perceive the regions of different scales

Simultaneously, different receptive fields can capture local information in different ranges. For example, one convolutional layer with 3 × 3 size can capture the information of 9 pixels near the calculation point, while double 3×3 convolution layers can capture the information within the range of 5×5. Thus, increasing the types of scales can better perceive the regions of different scales. Motivated by the above observations, we construct a multi-scale residual block based on the deep learning theory and the fact that the convolutional layers with different sizes can process different scales. Fig. 4(a) is a basic residual block composed of two 3×3 convolutional layers whose input and output are connected by identity mapping. Different from the basic residual block, Fig. 4(b) is a multi-scale residual block with two branches. One branch has identical structure with the basic residual block, while the other branch only has a single 3 × 3 convolutional layer. Both branches multiplied by different weights are fused by element-wise adding operation, that is, each branch makes different contributions to the output of the residual block.

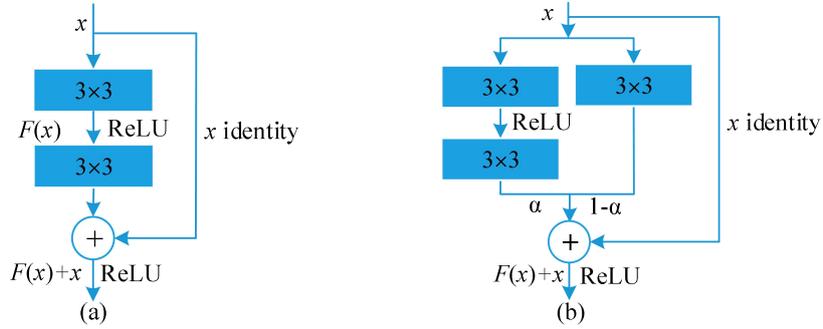

**Fig. 4** The basic residual block and the multi-scale residual block

The contribution of the left branch is greater than that of the right branch because the critical information of a sketch is still its outline which can be readily captured by the large receptive field. The output of a multi-scale residual block can be formulated as follows.

$$y = x + [\alpha \times F_{3,3}(x) + (1-\alpha) \times F_3(x)] \tag{6}$$

where $x$ represents the input of a multi-scale residual block, $y$ indicates its output, $F_3$ stands for a single convolutional layer, $F_{3,3}$ indicates double convolutional layers, and $\alpha$ is a weight. The expression in the bracket can be regarded as a residual function to be learned. Details of all multi-scale residual blocks are shown in Table 2.

It should be noted that formula (6) is used to compute the output of each multi-scale residual block in testing phase. If a multi-scale residual block is trained according to formula (6), we will not achieve a satisfactory model, because a multi-scale residual block is about 1.5 times of a basic residual block in the number of parameters. If element-wise adding is used to merge two different branches, the increase of parameters easily makes the network overfit, which undoubtedly increases the training difficulty. Therefore, we need to design a special training method. Specifically, we can activate one of the branches randomly as shown in Fig 5.

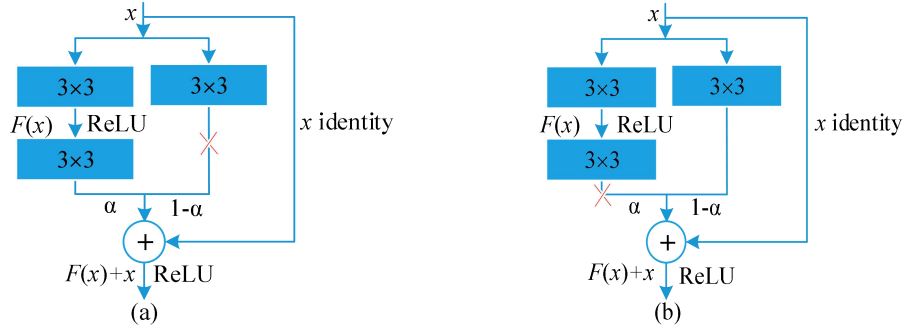

**Fig. 5** The left branch is activated in (a) and the right branch is activated in (b). A red cross indicates that the current branch is not activated

In training phrase, both branches of a multi-scale residual block do not work together, but one of the branches is activated randomly according to a probability obeying binomial distribution. Assuming that the network composed of basic residual blocks has $n_p$ parameters, so our network composed of multi-scale residual blocks has about $1.5 \times n_p$ parameters. The expectation of parameters using the activated randomly method is as follows.

$$E(n_{ar}) = \alpha \times n_p + (1-\alpha) \times \frac{n_p}{2}$$

$$= \frac{n_p(\alpha+1)}{2} \in [\frac{n_p}{2}, n_p] \qquad (7)$$

According to formula (7), the number of parameters in our network does not increase, but is less than the number of parameters in the network composed of basic residual blocks. The method not only prevents the overfitting of our network effectively but also reduces the negative impact of the poor gradient flow during the backward propagation, resulting in making different sketch features separate more obviously. We demonstrate experimentally the advantages of the multi-scale residual block compared with the basic residual block.

**3.3 Hierarchical residual structure**

The residual learning plays an important role in improving the network performance. ResNet uses residual learning in the building block for extending the depth of the network. However, it is difficult for this implementation to sufficiently learn the sketch features. DenseNet [37] adds a large number of shortcut connections which are not for residual learning but feature reuse in essence. Due to dense shortcut connections, it is not beneficial to enhance the sketch recognition accuracy. In order to make the network fully learn the sketch features and avoid adding unnecessary shortcut connections, this paper integrates residual learning into overall network structure. The proposed hierarchical residual structure is shown in the green dotted box of Fig. 2. The green dotted box is a recursive residual block. The innermost red blocks are multi-scale residual blocks with the structure of Fig. 5. Several multi-scale residual blocks are stacked to form a whole, and the input of its head and the output of its tail are added together by a shortcut connection to produce an inner residual block, as shown in the green blocks of Fig. 2. There are four inner residual blocks in the hierarchical residual structure. The four inner residual blocks are composed of 3, 4, 6 and 3 multi-scale residual blocks from left to right. Four inner residual blocks are stacked together as a whole, and the input of its head and the output of its tail are connected to form an outer residual block, as shown in the green dotted box. Multi-scale residual block, inner residual block and outer residual block constitute the hierarchical residual structure. Details of the hierarchical residual structure are shown in Table 2. There are two reasons why we make a residual structure with multiple levels. From the perspective of forward propagation, the outer residual learning enhances the influence of the useful shallow features on the deep features, and the inner residual learning enhances the influence of the useful features of the front inner residual blocks on the later inner residual blocks. From the perspective of backward propagation, the outer shortcut connection makes gradients directly flow to the shallow layers across multiple layers, avoiding the vanishing gradient. The inner shortcut connections make each residual block benefit from the gradients at the end of the network as much as possible, and improve the strength of gradients.

Moreover, the input dimensionality of the inner residual blocks and the outer residual block differ from their output dimensionality, it is necessary to reduce spatial dimensionality and increase channel dimensionality for short connections. Parameters of convolutional layers on the shortcut connections are shown in Table 3. It can be seen from Table 3 that we use the convolutional method instead of the pooling method for dimensionality reduction or increase. This is due to two reasons: (i) The convolutional method can realize dimensionality variation in both space and channel, while the pooling method can only achieve dimensionality reduction in space. (ii) The pooling method will filter out a lot of useful information in the condition of large span, which is not conducive to residual learning.

For the outermost shortcut connection, although the sketch features after dimensionality reduction can meet the requirements of residual learning, the conventional element-wise addition will increase training error and achieve an uncompetitive result. This phenomenon will not be eliminated by adjusting the learning rate and normalization. We found that scaling the information on the outer shortcut connection is helpful to solve the above problem. The information is multiplied by $\beta$ before it flows into the next layer. We suppose that the reason for scaling is that the effect of the trunk network to the overall network is the largest and the contribution of the shallow features to the final features is limited due to the large span.

**Table 3** Parameters on the shortcut connections of the inner and outer residual blocks

| Residual block | Kernel | Stride | Padding | Input dimensionality | Output dimensionality |
| --- | --- | --- | --- | --- | --- |
| Outer residual block | $9\times 9$ | $8\times 8$ | 1 | 64 | 512 |
| First inner residual block | $1\times 1$ | $1\times 1$ | 0 | 64 | 64 |
| Second inner residual block | $1\times 1$ | $2\times 2$ | 0 | 64 | 128 |
| Third inner residual block | $1\times 1$ | $2\times 2$ | 0 | 128 | 256 |
| Fourth inner residual block | $1\times 1$ | $2\times 2$ | 0 | 256 | 512 |

Multi-scale residual blocks exploit residual learning to establish a hierarchical residual structure which makes our network have better learning ability. In the experiments, we validate the effect of inner residual learning and outer residual learning on performance.

### 3.4 Compact triplet-center loss

Although the triplet-center loss aims at minimizing intra-class distances and maximizing inter-class distances, it cannot fully compress too large intra-class distances of sketches and enlarge too small inter-class distances of sketches. We show two kinds of distances in Fig. 6 when the triplet-center loss is exploited for our network. One is the distances from the sketch features to their corresponding centers, and the other is the distances from the sketch features to their negative centers (the corresponding centers of other classes). As you can see, the distances from the sketch features to their corresponding centers are approximately equal to the distances from the sketch features to their negative centers. Therefore, in order to enhance the distinction of the sketch features, we need to make great efforts in compressing the intra-class distances and keeping the reasonable inter-class distances. For simplicity, $D_{t1}$ stands for the intra-class distances and $D_{t2}$ stands for the inter-class distances when the triplet-center loss is used. Theoretically, it is possible for some neural networks to enhance their discriminative power for the sketch features through heavily increasing the hyper-parameter $m$ in the triplet-center loss. However, $D_{t1}$ and $D_{t2}$ will become larger simultaneously when the hyper-parameter $m$ is greatly increased.

The reason for the above phenomenon is that the triplet-center loss cannot restrict the update direction of $D_{t1}$, that is, $D_{t1}$ may become larger or smaller. Therefore, it is very important to restrict the update direction of $D_{t1}$ and $D_{t2}$. We define a compact triplet-center loss which can fully compress the intra-class distances and keep the reasonable inter-class distances by restricting the update directions. The specific definition is as follows.

$$L_{ct} = \sum_{i=1}^{M} \max[m \times D(x_i - c_{y_i}) - D(x_i - c_{y_{j,j \neq i, j \in random}}), 0] \qquad (8)$$

where $x_i \in R^d$ represents the $i$th sketch features usually from neurons in the previous layer of the loss. $c_{y_i} \in R^d$ indicates the center of the corresponding category $y_i$. $c_{y_j} \in R^d$ is a randomly selected negative center. $D(.)$ stands for the squared Euclidean distance. Similarly, for simplicity, $D_{ct1}$ stands for the intra-class distances and $D_{ct2}$ stands for the inter-class distances, when we use the proposed loss. Obviously, $D_{ct1}$ can only get smaller, because $D_{ct1}$ updates faster than $D_{ct2}$ and the proposed loss can only be reduced during training. Simultaneously, the compact triplet-center loss can keep the ratios of the intra-class distances to the inter-class distances suitable, which ensures the distinctions between all classes. There are three advantages to multiplying m by the distances from the sketch features to their corresponding centers. First, the intra-class distances can be fully compressed as long as they are larger than $1/m$ of the distances from the sketch features to their negative centers. Second, the proposed network can converge readily when $m$ times of the distances from the sketch features to their corresponding centers are roughly equal to the distances from the sketch features to their negative centers. Third, it is convenient to determine the value of the hyper-parameter $m$ in the loss. In addition, we change the strategy of selecting the negative centers, that is, we randomly select the negative centers instead of selecting the nearest negative centers to the samples, which can further improve the recognition accuracy. The reason for improvement is that the random strategy can break the balance condition for convergence and lead to further compression of the distances between identical categories to the minimum.

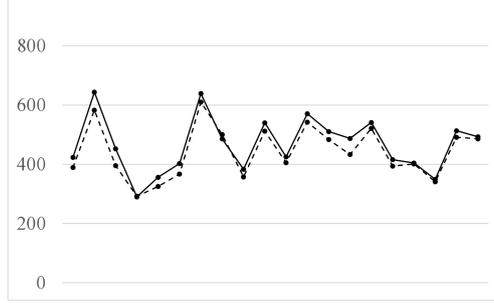

**Fig. 6** The horizontal axis represents 20 samples from different classes. The vertical axis represents the distances between the samples and their corresponding or negative centers. The solid line indicates the distances between the samples and their negative centers, and the dotted line indicates the distances between the samples and their corresponding centers

The key of the compact triplet-center loss is to update all class centers. According to formula (8), we use the method similar to [36] to update all class centers. The formula for updating the class $k$ can be defined as

$$\Delta c_k = \sum_{i=1}^{n}[\eta m \varphi(y_i = k)(x_i - c_{y_i}) + \eta \varphi(y_j = k)(c_{y_j} - x_i)] \qquad (9)$$

where $\Delta c_k$ represents the update of the center of class $k$, $\eta \in [0,1]$ is the learning rate. The value of $\varphi(.)$ is 1 or 0. When the expression in the brackets of $\varphi(.)$ is equal to true, 1 is taken, or 0 is taken on the contrary. The update of the center of class $k$ comes from two parts: one is from the differences between the sketch features with class $k$ and the centers of the class $k$. The other is from the differences between the centers of the selected randomly class $k$ and the sketch features. Similarly, the update of $x_i$ can be derived as follows.

$$\Delta x_i = \eta[(m-1)x_i - mc_{y_i} + c_{y_j}] \qquad (10)$$

From the above discussion, we can conclude that the compact triplet-center loss is differentiable, so it is easy to train the neural network by using the stochastic gradient descent method. Compared with the triplet-center loss, the compact triplet-center loss can perform the intra-class compression thoroughly. Therefore, the sketch features are more discriminative when the compact triplet-center loss is used as the supervised signal which is also the key of sketch recognition.

## 4 Experiment

### 4.1 Dataset and settings

#### 4.1.1 Dataset

Tu-Berlin [10] is the first large-scale sketch dataset. The dataset consists of 250 categories that often appear in daily life, with a total of 20000 samples drawn by 1350 volunteers via crowdsourcing. The size of each sample is $1111 \times 1111$. We scale the samples to $255 \times 255$. At present, this dataset has been widely used in the sketch research and the majority of sketch recognition experiments.

### 4.1.2 Regular data augmentation

There are 20000 samples in Tu-Berlin, which is far less than the number of samples in a natural image dataset. Consequently, we adopt an appropriate data augmentation technology. Specifically, for any input sample, we rotate it -5~5 degrees to left or right, flip it horizontally, and move 0 ~ 31 pixels to the right or down or simultaneously, then cut to 224×224 size, so that the number of samples increases to 11×32×32×2 times of the original dataset.

### 4.1.3 Settings

We use the PyTorch framework to code the proposed network and use the Adam algorithm to train it on a GTX 1080 GPU. All networks will be trained for 180 epochs and the initial learning rate is 0.001. The learning rate is reduced to 0.65 times of the original learning rate every 10 epochs. In order to prevent the learning rate from being too small to continue learning, it is updated to 0.95 times of the original learning rate every 20 epochs when the epoch is larger than 100. In addition, the batch size is 28 and the initial weights are assigned by Kaiming uniform distribution. We use the 3-fold cross validation method to train and test, that is, divide the samples into 3 splits, 2/3 for training and 1/3 for testing. In addition, we randomly select 15% training samples as the validation dataset and use this dataset to determine the optimal hyper-parameters based on our experience. In the experiment, $\lambda$ is set to 0.024.

### 4.1.4 Competitors

The baseline models are mainly classified into three categories. The first category is based on the hand-crafted feature methods including HOG-SVM [10], Fisher Vector [11] and multi-kernel learning [12]. The second category is based on the deep learning methods including Sketch-a-Net [1], Sketch-BERT [2], SketchNet [13], transfer learning [14]. The last category is the image classification networks including VGGNet [21], GoogLeNet [22], ResNet-34 [19], ResNet-50 [19], EfficientNet [28], DenseNet [37] and BoTNet [38].

### 4.2 Experimental results

#### 4.2.1 Comparative results

We draw a table about the recognition results as follows.

In Table 4, the proposed hierarchical residual network outperforms most of baseline methods including the hand-crafted methods: HOG-SVM, Fisher Vector and Multi-kernel learning; the deep learning methods: SketchNet, Transfer learning; the deep learning methods for image classification: VGGNet, GoogLeNet, ResNet-34, ResNet-50, DenseNet, EfficientNet, BoTNet. These results prove that our method is efficient. The performance of the proposed method is better than the above methods because our network can acquire the discriminative sketch features by three components. Firstly, multi-scale perception conforms to the human vision characteristics which are rarely considered in the above methods. Secondly, the proposed method is based on a hierarchical structure which greatly reduces the possibility of vanishing gradient. Finally, the loss in our method is the compact triplet-center loss which has the ability to compress too large intra-class space in the sketch field, while many other methods only use the cross entropy loss which does not have this ability. Based on the above three components, the distinguishability of the sketch features can be greatly improved, resulting in the high recognition accuracy of the network.

Table 4 The recognition results of all models on Tu-Berlin. Sketch-a-Net is a four-branch ensemble. The computational complexity of Sketch-BERT is positively related to the number of sketch strokes. Here, the average number of sketch strokes is taken.

| Model | Accuracy | FLOPs | Parameters |
| --- | --- | --- | --- |
| HOG-SVM | 56% | - | - |
| Fisher vector | 68.9% | - | - |
| Multi kernel | 65.8% | - | - |
| Sketch-a-Net | 77.95% | 3.0G | 34.1M |
| Sketch-BERT | 76.3% | 4.4G | 85.6M |
| SketchNet | 75.95% | 2.3G | 63.5M |
| Transfer learning | 72.5% | 2.7G | 65.6M |
| VGGNet | 72.85% | 11.3G | 130M |
| GoogLeNet | 71.25% | 1.6G | 6.2M |
| ResNet-34 | 72.31% | 3.7G | 21.4M |
| ResNet-50 | 71.08% | 4.1G | 24M |
| DenseNet | 70.5% | 4.3G | 18.6M |
| EfficientNet | 72.65% | 0.4G | 4.3M |
| BoTNet | 70.28% | 4.1G | 19.3M |
| humans | 73.1% | - | - |
| **Our Model** | **76.14%** | **3.8G** | **34M** |

In terms of computational complexity, our network is comparable to or better than some networks including VGGNet, ResNet, DenseNet, BoTNet, but the recognition accuracy of our network is much higher than these networks. In terms of the number of parameters, the number of parameters of our network is less than some networks including SketchNet, Transfer learning, VGGNet, and the recognition accuracy of our network is higher than them.

Although our method performance is worse than Sketch-BERT and Sketch-a-Net, our method still has advantages and significance. Both Sketch-BERT and Sketch-a-Net usually rely on stroke sequences to improve the quality of the sketch features. However, it is difficult to obtain stroke sequences for the datasets consisted of static sketch images. Both networks need to be pre-trained by using a large number of the samples, which increases the training costs and enhance data requirements. If they are not pre-trained, the performance of these networks will be greatly reduced. In addition, the proposed network has a more convenient and concise training process than Sketch-BERT, Sketch-a-Net and SketchNet. Overall, our network is a network with ordinary performance in terms of the computational and space complexity, but its recognition ability exceeds most networks and it has more excellent properties.

Moreover, the accuracy of all hand-crafted methods is lower than that of the deep learning methods. The latter is close to or higher than 73.1% of the human level, while our method is 3.04% higher than the human level. The performance of ResNet-34 is better than that of ResNet-50 indicating that the sketch recognition accuracy has nothing to do with the depth. Although the residual learning is used in ResNet-34 and ResNet-50, it is obvious that our network is better than them. Furthermore, DenseNet is far behind our network, which shows that too dense shortcut connections are not conducive to sketch feature learning.

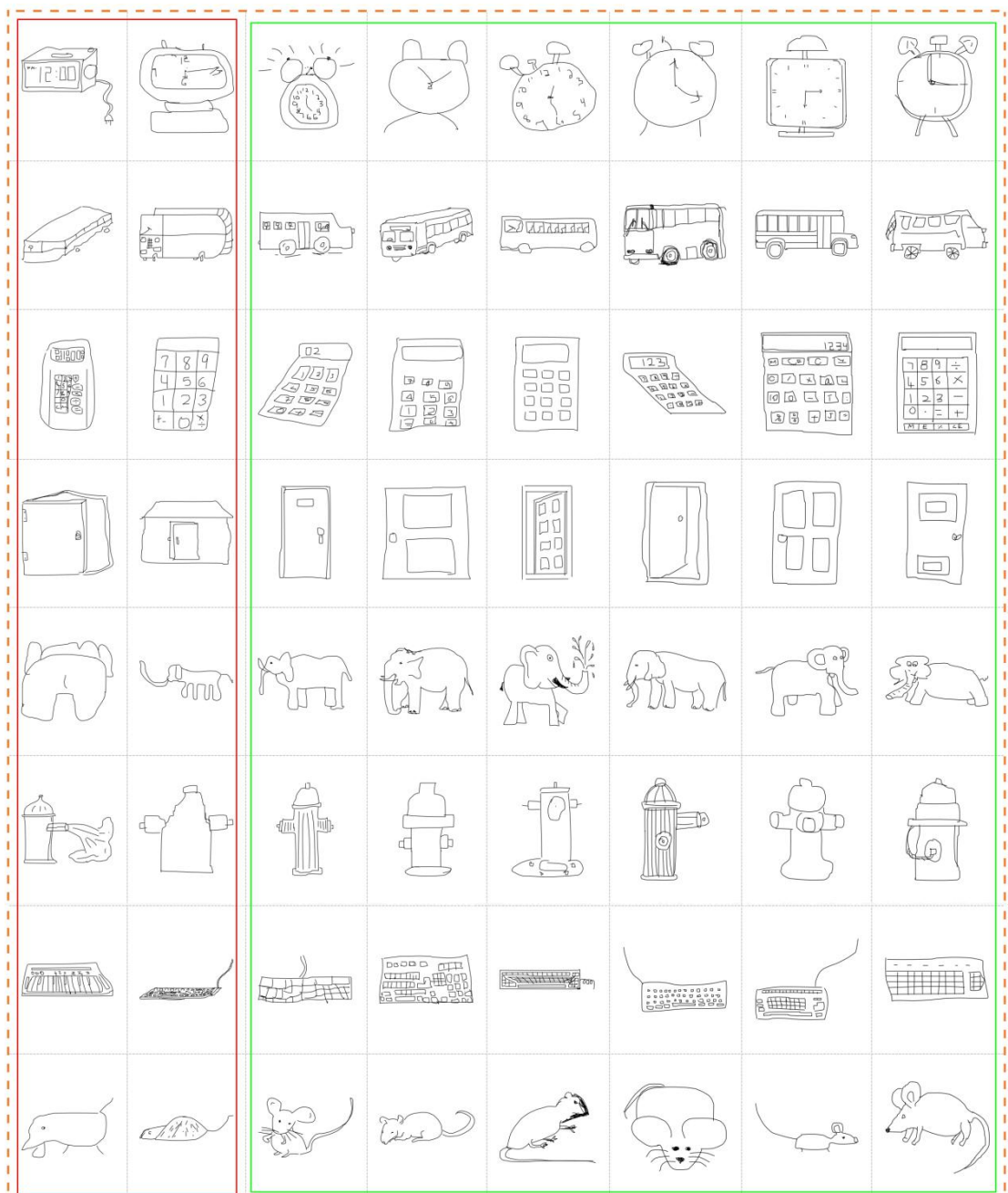

**Fig. 7** The qualitative results of our proposed network. The sketches in the red box are the samples that was recognized incorrectly, while the sketches in the green box are the samples that was recognized correctly. It is understandable that the incorrectly recognized samples are usually caused by special viewing angles, poor drawing levels, redundant objects and so on.

### 4.2.2 Qualitative results

Some qualitative experiment results are presented in Fig. 7. It can be seen that humans cannot usually recognize the sketches that are incorrectly recognized by our network. The sketches that are recognized incorrectly usually have the following characteristics: (i) The painting level is very poor. The related examples lie in the second column of the fifth row and the first two column of the last row. (ii) There are great differences in the viewing angles. The corresponding sketches include the first elephant in the fifth row and the second keyboard in the seventh row. (iii) There are redundant objects. For example, the first sketch in the sixth row is a fire hydrant with water, and the second sketch in the fourth row is a house with a door. (iv) The sketches belong to different sub categories. For example, the first sketch in the first row belongs to electronic watch which is different from the alarm clocks shown in the identical row.

### 4.3 Ablation analysis

The performance improvement of the proposed network mainly comes from two changes including the network structure changes and the compact triplet-center loss. To be specific, the network changes can improve the recognition accuracy by 2.22% while the compact triple-center loss can improve the recognition accuracy by 1.61%. More details can be found as follows.

### 4.3.1 Multi-scale residual block

In order to verify the effectiveness of multi-scale residual blocks, we replace all multi-scale residual blocks in our proposed network with basic residual blocks. The recognition results of our network with multi-scale residual blocks and the network with basic residual blocks are drawn as follows.

**Table 5** The performance comparison between the basic residual block and multi-scale residual block.

| Types of the innermost residual blocks | Accuracy(%) |
|---|---|
| Basic residual block | 75.28 |
| Multi-scale residual block | 76.14 |

In Table 5, the recognition result of our network with multi-scale residual blocks is obviously better than that of the network with basic residual blocks. The accuracy of the former is 0.86% higher than that of the latter, so the multi-scale residual block is obviously effective in improving the sketch features.

### 4.3.2 Inner residual learning and outer residual learning

We focus on the performance of the inner residual block and the outer residual block. Table 6 shows three cases that the network with inner residual learning, with outer residual learning and with both kinds of residual learning.

**Table 6** The influence of outer residual learning and inner residual learning. Multi-scale residual blocks are used by default in the following three cases.

| Networks with different residual blocks | Accuracy(%) |
|---|---|
| The network with inner residual block | 75.26 |
| The network with outer residual block | 75.76 |
| The network with inner and outer residual block | 76.14 |

It is observed that both inner residual learning and outer residual learning can improve the performance of the network. Specifically, the outer residual learning improves the accuracy by 0.88%, and the inner residual learning improves the accuracy by 0.38%. The effect of outer residual learning is stronger than that of inner residual learning. The reason for this phenomenon is that the outer residual learning can solve the problem of more serious gradient vanishment caused by multiple stacked layers. In addition, the information on the shortcut connection used for outer residual learning does not directly flow into the trunk network. We perform compression for the information by adjusting the hyper-parameter $\beta$, and draw a figure about accuracy using different $\beta$ as follows.

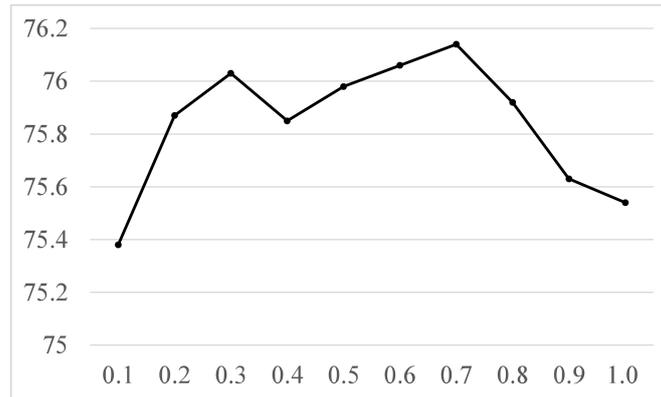

**Fig. 8** The horizontal axis represents the hyper-parameter $\beta$. The vertical axis represents the recognition accuracy

From Fig. 8, different hyper-parameters $\beta$ can obtain different accuracy. The accuracy essentially increases with the increase of $\beta$, and reaches the maximum when $\beta$ is equal to 0.7, then decreases with the increase of $\beta$. For hyper-parameter $\beta$, too large or too small values both result in undesirable results. When $\beta$ is too small, the shallow features will have little influence on the deep features. When $\beta$ is too large, some useless shallow features will influence the deep features. Therefore, it is very important to choose a correct value.

**4.3.3 Compact triplet-center loss**

To evaluate the superiority and effectiveness of the proposed loss, we connect the triplet-center loss and the compact triplet-center loss to the end of the hierarchical residual network and use these two losses to guide the network training respectively. *m* is set to 4.5 in the proposed loss and *m* is set to 5.0 in the original loss respectively because their best recognition accuracy are achieved using these two values. The results of the comparative experiment are shown in Table 7.

**Table 7** The performance comparison between the triplet-center loss and our proposed loss.

| Loss types | Accuracy(%) |
|---|---|
| Triplet-center loss | 75.07 |
| Compact triplet-center loss | 76.14 |

In Table 7, the accuracy of the compact triplet-center loss is 76.14% while the accuracy of the triplet-center loss is 75.07%. The proposed loss evidently outperforms the triplet-center loss. In addition, we randomly choose 25 classes to display the distributions of two kinds of features produced by these two losses using t-SNE [39] technology. As is shown in Fig. 9, the sketch features in Fig. 9(a) are more compact than the sketch features in Fig. 9(b), and the number of the sketch features far from their corresponding centers in Fig. 9(a) is obviously less than the number in Fig. 9(b), which verifies the ability for stronger compression in our proposed loss.

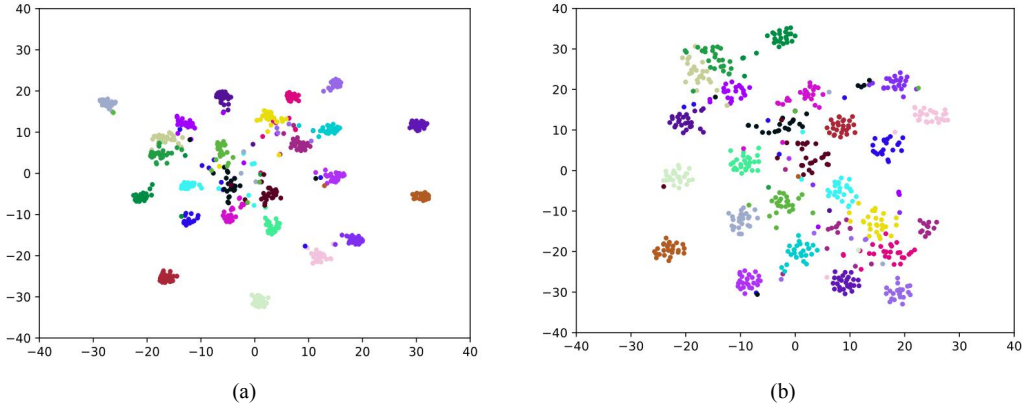

**Fig. 9** (a) is the distribution of the sketch features using the compact triplet-center loss. (b) is the distribution of the sketch features using the triplet-center loss. Different colors indicate different classes

In addition, in order to verify the necessity of the proposed loss, that is, the fact that we cannot make the triplet-center loss suitable for sketch recognition by adjusting the hyper-parameter $m$, we conduct adequate experiments using different hyper-parameter $m$ (5, 10, 25, 50, 100, 200, 300, 400, 500, 1000), and some results are shown in Fig. 10 and Fig 11.

Evidently, the margins between two adjacent classes are increased indeed, while the intra-class distances and their fluctuating ranges are also enlarged, resulting in lower recognition accuracy. From Fig. 10 and Fig. 11, we can conclude that it is more important for the sketch recognition task to decrease the ratio of intra-class distances to their corresponding margins and their fluctuating ranges. This is solved by the proposed loss.

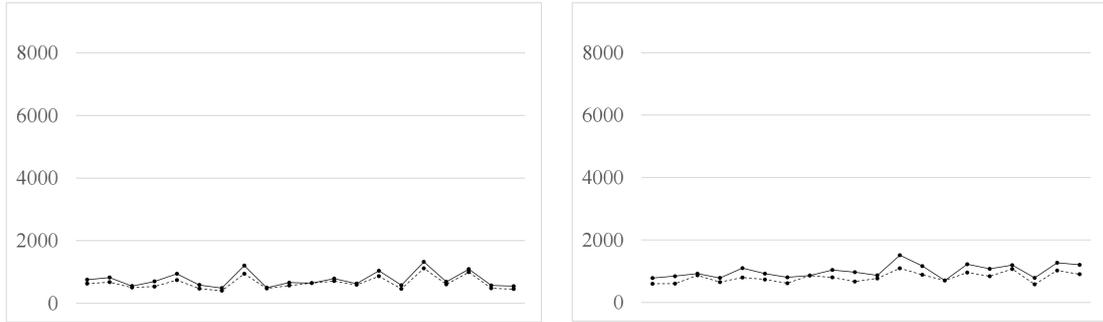

**Fig. 10** The horizontal axis represents 20 samples from different classes. The vertical axis represents the distances between the samples and their related centers. The solid line indicates the distances between the samples and their negative centers, and the dotted line indicates the distances between the samples and their corresponding centers. The left figure uses the network with the hyper-parameter *m* **50**, and its recognition accuracy is 75.07%. The right figure uses the network with the hyper-parameter *m* **100**, and its recognition accuracy is 74.8%

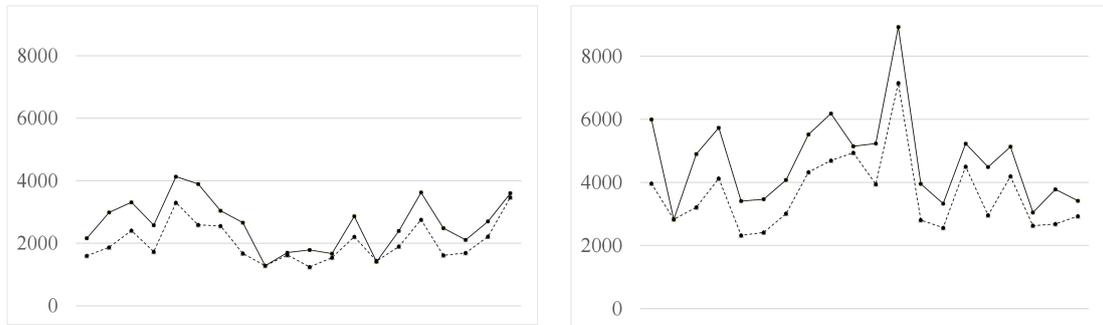

**Fig. 11** The horizontal axis represents 20 samples from different classes. The vertical axis represents the distances between the samples and their related centers. The solid line indicates the distances between the samples and their negative centers, and the dotted line indicates the distances between the samples and their corresponding centers. The left figure uses the network with the hyper-parameter *m* **300**, and its recognition accuracy is 74.36%. The right figure uses the network with the hyper-parameter *m* **500**, and its recognition accuracy is 74.25%

### 4.3.4 Hyper-parameter *m*

The hyper-parameter *m* controls compression intensity of the distances from the sketch features to their corresponding centers and the dispersion level of different classes. The value of *m* plays an important role in classification. Therefore, we evaluate the influence of the hyper-parameter *m* on our network. The results of different *m* are shown in Fig. 12.

Obviously, it is the optimal choice when *m* is set to 4.5. When the hyper-parameter m is less than 2.0, the sketch features are too close to their negative centers to be better separated from different classes. When the hyper-parameter is greater than 5.5, the network is difficult to be trained and cannot achieve the optimal accuracy. Therefore, the right *m* will optimize the deep features. Simultaneously, we can observe that the network performance is relatively stable in the certain range of $m \in [2.5, 5.5]$.

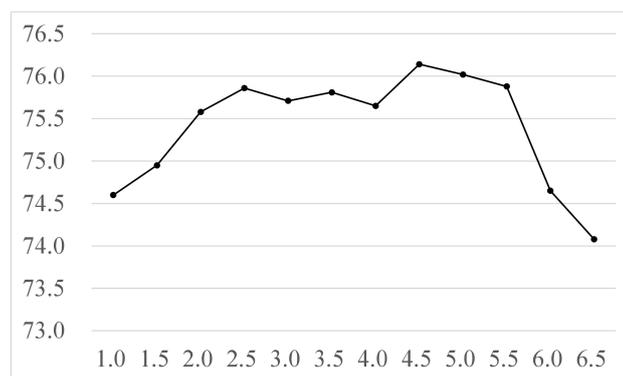

**Fig. 12** The horizontal axis represents the hyper-parameter *m*. The vertical axis represents the recognition accuracy

## 5 Conclusion

We propose a hierarchical residual network for sketch recognition. The network achieves 76.14% of the accuracy, and it beats most of baseline models. Meanwhile, without specific data augmentation, our network has excellent performance in all non-sequential models. The high performance of the proposed network results from three key points including multi-scale residual block, hierarchical residual structure and compact triplet-center loss. The multi-scale residual block solves the learning problem of the sketch features with different scales. The hierarchical residual structure strengthens the flow of gradients in our network and reduces training error. The compact triplet-center loss is exploited to solve the problem of too large intra-class distances and too small inter-class distances. As is known to all, there are a large number of sketch samples on the Internet. These samples can be used to improve the recognition capability of our network, but they usually lack manual labels. Therefore, we will focus on the unsupervised learning methods for sketch recognition in the future. In addition, it is worth noting that the proposed method not only can be used for sketch recognition, but also can be integrated into the fields of sketch-based image retrieval and sketch-based 3D image retrieval which are also our research work in the future.

**Acknowledgements**  This work was supported partly by the National Natural Science Foundation of China (No. 61379065) and the Natural Science Foundation of Hebei province in China (No. F2019203285).

**References**

[1] Yu Q, Yang Y, Liu F, et al. (2017). Sketch-a-Net: a deep neural network that beats humans[J]. International Journal of Computer Vision 122(3):411-425.
[2] Lin H, Fu Y, Jiang Y G, et al. (2020) Sketch-BERT: learning sketch bidirectional encoder representation from transformers by self-supervised learning of sketch gestalt[C]. In: IEEE Conference on Computer Vision and Pattern Recognition. Seattle, WA, USA, pp 6757-6766.
[3] Klare B, Li Z, Jain A K. (2011) Matching forensic sketches to mug shot photos[J]. IEEE Trans on Pattern Analysis and Machine Intelligence 33(3):639-646.
[4] Ouyang S, Hospedales T M, Song Y Z, et al. (2020) Forgetmenot: memory-aware forensic facial sketch matching[C]. In IEEE Conference on Computer Vision and Pattern Recognition. Seattle, WA, USA, pp 5571-5579.

[5] Pang K, Li K, Yang Y, et al. (2019) Generalising fine-grained sketch-based image retrieval[C]. In: IEEE Conference on Computer Vision and Pattern Recognition. Long Beach, CA, USA, pp 677-686.

[6] Yu Q, Liu F, Song Y Z, et al. (2016) Sketch me that shoe[C]. In: IEEE Conference on Computer Vision and Pattern Recognition. Salt Lake City, UT, USA, pp 799-807

[7] Pang K, Yang Y, Hospedales T M, et al. (2020) Solving mixed-modal jigsaw puzzle for fine-grained sketch-based image retrieval[C]. In: IEEE Conference on Computer Vision and Pattern Recognition. Seattle, WA, USA, pp 10344-10352.

[8] Wang F, Lang L, Li Y. (2015) Sketch-based 3d shape retrieval using convolutional neural networks[C]. In: IEEE Conference on Computer Vision and Pattern Recognition. Boston, MA, USA, pp 1875-1883.

[9] Chen J, Qin J, Liu L, et al. (2019) Deep sketch-shape hashing with segmented 3D stochastic viewing[C]. In: IEEE Conference on Computer Vision and Pattern Recognition. Long Beach, CA, USA, pp 791-800.

[10] Eitz M, Haysy J, Alexa M. (2012) How do humans sketch objects?[J]. ACM Transactions on Graphics 31(4):1-10.

[11] Schneider R G, Tuytelaarsy T. (2014) Sketch classification and classification-driven analysis using fisher vectors[J]. ACM Transactions on Graphics 33(6):174.1-174.9.

[12] Li Y, Hospedales T M, Song Y Z, et al. (2015) Free-hand sketch recognition by multi-kernel feature learning[J]. Computer Vision and Image Understanding 137:1-11.

[13] Zhang H, Liu S, Zhang C, et al. (2016) SketchNet: Sketch classification with web images[C]. In: IEEE Conference on Computer Vision and Pattern Recognition. Las Vegas, NV, USA, pp 1105-1113.

[14] Sert M, Boyaci E. (2019) Sketch recognition using transfer learning[J]. Multimedia Tools and Applications 78:17095-17112.

[15] Yang L, Sain A, Li L P, et al. (2020) $S^3$net: Graph representational network for sketch recognition[C]. In: IEEE International Conference on Multimedia and Expo. London, United kingdom, pp 1-6.

[16] Xu P, Huang Y Y, Yuan T T, et al. (2018) SketchMate: deep hashing for million-scale human sketch retrieval[C]. In: IEEE Conference on Computer Vision and Pattern Recognition. Salt Lake City, UT, USA, pp 8090-8098.

[17] He J Y, Wu X, Jiang Y G, et al. (2017) Sketch recognition with deep visual-sequential fusion model[C]. In: Proceedings of the ACM Multimedia Conference. Mountain View, CA, USA, pp 448-456.

[18] Zheng Y, Yao H X, Sun X S, et al. (2021) Sketch-specific data augmentation for freehand sketch recognition[J]. Neurocomputing https://doi.org/10.1016/j.neucom.2020.05.124.

[19] He K M, Zhang X Y, Ren S Q, et al. (2016) Deep residual learning for image recognition. In: IEEE Conference on Computer Vision and Pattern Recognition. Las Vegas, NV, USA, pp 770–778.

[20] Krizhevsky A, Sutskever I, and Hinton G E. (2012) Imagenet classification with deep convolutional neural networks[C]. In: Annual Conference on Neural Information Processing Systems. Lake Tahoe, NV, USA, pp 1106-1114.

[21] Simonyan K, Zisserman A. (2015) Very deep convolutional networks for large-scale image recognition[C]. In: International Conference on Learning Representations. San Diego, CA, USA, pp 1-14.

[22] Szegedy A, Liu W, Jia Y, et al. (2015) Going deeper with convolutions[C]. In: Proceedings of the

IEEE Conference on Computer Vision and Pattern Recognition. Boston, MA, USA, pp 1-9.

[23] Zhang L, Jiao L C, Ma W P, et al. (2020) PolSAR image classification based on multi-scale stacked sparse autoencoder[J]. Neurocomputing 351:167-179.

[24] Huang G, Chen D L, Li T H, et al. (2018) Multi-scale dense networks for resource efficient image classification[C]. In: International Conference on Learning Representations. Vancouver, BC, Canada, pp 4700-4708.

[25] Pang Y, Zhao X, Zhang L, et al. (2020) Multi-scale interactive network for salient object detection[C]. In: Proceedings of the IEEE Computer Society Conference on Computer Vision and Pattern Recognition. Seattle, WA, USA, pp 9410-9419.

[26] Zhao T, Wu X. (2019) Pyramid feature attention network for saliency detection[C]. In: IEEE Computer Society Conference on Computer Vision and Pattern Recognition. Long Beach, CA, USA, pp 3080-3089.

[27] Zhao H, Shi J, Qi X, et al. (2017) Pyramid scene parsing network[C]. In: Proceedings of IEEE Conference on Computer Vision and Pattern Recognition. Honolulu, HI, USA, pp 6230-6239.

[28] Tan M X, Le Q V. (2019) EfficientNet: Rethinking Model Scaling for Convolutional Neural Networks[C]. In: International Conference on Machine Learning. Long Beach, CA, United states, pp 10691-10700.

[29] Zagoruyko S, Komodakis N. (2016) Wide residual networks[C]. In: British Machine Vision Conference. UK, 87.1-87.12.

[30] Yeh C H, Huang C H, Kang L W. Multi-scale deep residual learning-based single image haze removal via image decomposition[J]. IEEE Transactions on Image Processing 29(12): 3153-3167.

[31] Ju D, Zhang P P, Wang D, et al. (2019) Video person re-identification by temporal residual learning[J]. IEEE Transactions on Image Processing 28(3):1366-1377.

[32] Wang F, Jiang M, Qian C, et al. (2017) Residual attention network for Image classification[C]. In: IEEE Conference on Computer Vision and Pattern Recognition. Honolulu, HI, USA, pp 6450-6458.

[33] Wang M, Deng W. (2020) Mitigating bias in face recognition using skewness-aware reinforcement learning[C]. In: Proceedings of the IEEE Computer Society Conference on Computer Vision and Pattern Recognition. Seattle, WA, USA, pp 9319-9328.

[34] Chang J, Lan Z, Cheng C, et al. (2020) Data uncertainty learning in face recognition[C]. In: IEEE Conference on Computer Vision and Pattern Recognition. Seattle, WA, USA, pp 5709-5718.

[35] Qiu H Q, Li H L, Wu Q B, et al. (2020) Offset bin classification network for accurate object detection[C]. In: IEEE Conference on Computer Vision and Pattern Recognition. Seattle, WA, USA, pp 13185-13194.

[36] He X, Zhou Y, Zhou Z, et al. (2018) Triplet-center loss for multi-view 3D object Retrieval[C]. In: IEEE Conference on Computer Vision and Pattern Recognition. Salt Lake City, UT, USA, pp 1945-1954.

[37] Huang G, Liu Z, Van Der Maaten L, et al. (2017) Densely connected convolutional networks[C]. In: IEEE Conference on Computer Vision and Pattern Recognition. Long Beach, CA, USA, pp 2261-2269.

[38] Srinivas A, Lin T Y, Parmar N K, et al. (2021) Bottleneck Transformers for Visual Recognition. https://arxiv.org/abs/2101.11605.

[39] Laurens V D M, Hinton G. (2008) Visualizing Data using t-SNE[J]. Journal of Machine Learning Research 9:2579-2605.